\documentclass[10pt,twocolumn,letterpaper]{article}

\usepackage[pagenumbers]{cvpr} 

\usepackage{graphicx}
\usepackage{amsmath}
\usepackage{amssymb}
\usepackage{booktabs}
\usepackage{algorithm}
\usepackage{algorithmic}
\usepackage{multirow}
\usepackage{makecell}
\usepackage{array}
\usepackage{multicol}
\usepackage{enumitem}
\usepackage{balance}
\usepackage[pagebackref,breaklinks,colorlinks]{hyperref}

\usepackage[capitalize]{cleveref}
\crefname{section}{Sec.}{Secs.}
\Crefname{section}{Section}{Sections}
\Crefname{table}{Table}{Tables}
\crefname{table}{Tab.}{Tabs.}



\begin{document}


\title{Visualizing and Understanding Patch Interactions in Vision Transformer}




\author{
 Jie Ma$^{1,2,3*\dagger}$~~ Yalong Bai$^{3\dagger}$~~ Bineng Zhong$^{1\S}$~~ Wei Zhang$^3$~~ Ting Yao$^3$~~ Tao Mei$^3$ \\
 {\normalsize $^1$Guangxi Normal University~~ \normalsize $^2$Huaqiao University~~  \normalsize $^3$JD Explore Academy}\\
 \tt\small majie@stu.hqu.edu.cn~~ ylbai@outlook.com~~ bnzhong@gxnu.edu.cn \\ \tt\small \{wzhang.cu, tingyao.ustc\}@gmail.com~~ tmei@jd.com 
}

\setlength{\skip\footins}{7pt}

\newcommand\blfootnote[1]{ 
\begingroup 
\renewcommand\thefootnote{}{\footnote{#1}} 
\addtocounter{footnote}{-1}
\endgroup
}

\maketitle
\setlength{\abovedisplayskip}{2pt}
\setlength{\belowdisplayskip}{2pt}
\begin{abstract}

Vision Transformer (ViT) has become a leading tool in various computer vision tasks, owing to its unique self-attention mechanism that learns visual representations explicitly through cross-patch information interactions.
Despite having good success, the literature seldom explores the explainability of vision transformer, and there is no clear picture of how the attention mechanism with respect to the correlation across comprehensive patches will impact the performance and what is the further potential. 
In this work, we propose a novel explainable visualization approach to analyze and interpret the crucial attention interactions among patches for vision transformer.
Specifically, we first introduce a quantification indicator to measure the impact of patch interaction and verify such quantification on attention window design and indiscriminative patches removal. Then, we exploit the effective responsive field of each patch in ViT and devise a window-free transformer architecture accordingly. Extensive experiments on ImageNet demonstrate that the exquisitely designed quantitative method is shown able to facilitate ViT model learning, leading the top-1 accuracy by 4.28\% at most. Moreover, the results on downstream fine-grained recognition tasks further validate the generalization of our proposal. 

\end{abstract}
\blfootnote{$*$~This work was done at JD Explore Academy.}
\blfootnote{$\dagger$~Equal contribution.~~~$\S$ Corresponding author.}
\section{Introduction}

\begin{figure}
 \centering
    \includegraphics[width=0.85\linewidth,page=1]{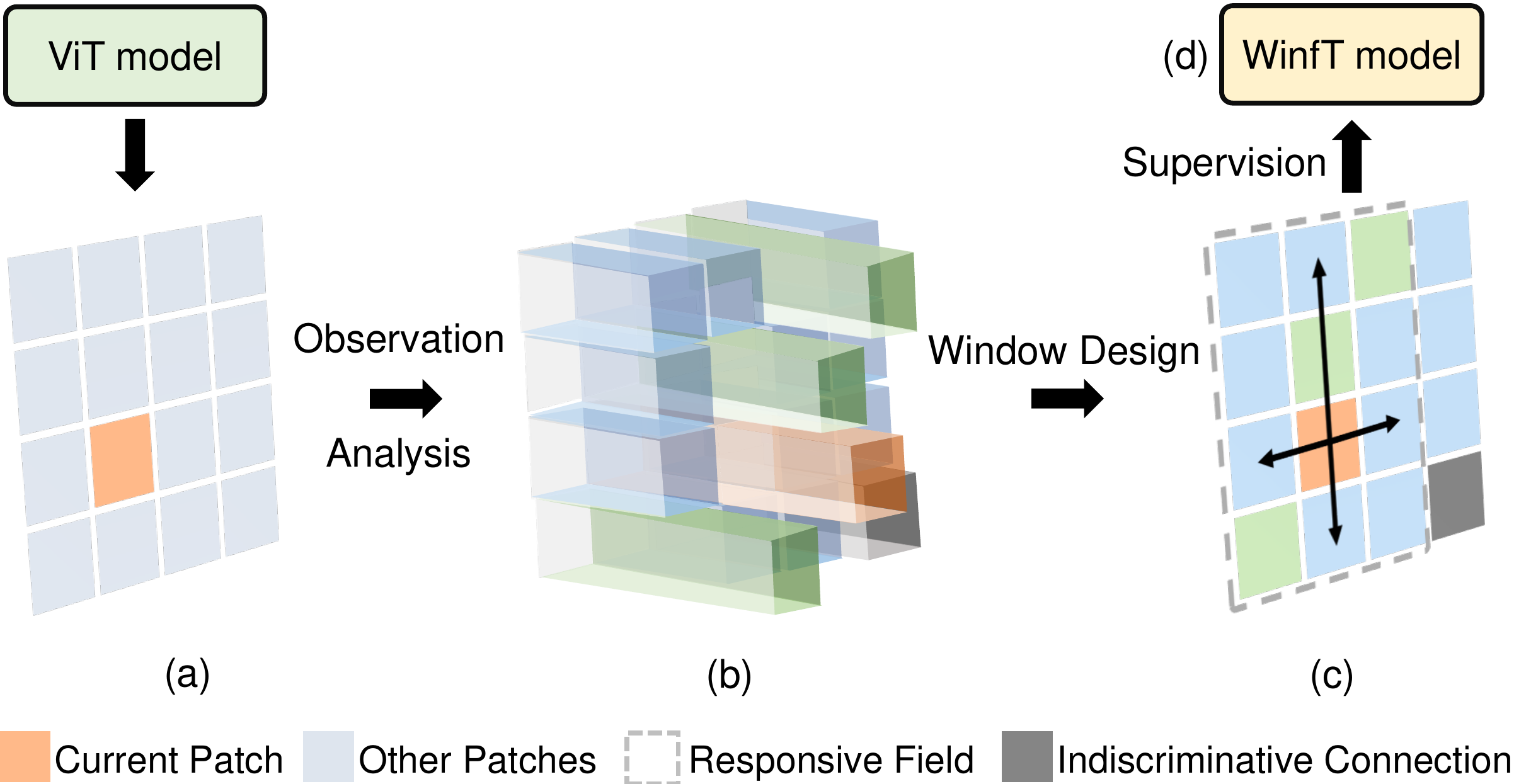}

 \caption{
   Given a patch (\textcolor[RGB]{255,185,137}{yellow}) in (a), we first analyze and quantify the impact of interactions between the current patch and all other patches. According to our patch interaction quantitative results, 
   we observe the distributions of relevant patches for current patch in (b), figure out the boundary of responsive field that containing patches with critical information to current patch in (c).
   Then, we apply the responsive field for guiding vision transformer model training. The performance improvement from ViT to WinfT in (d) can be regarded as a strong posteriori proof of the rationality of our proposed explainable visualization schema for ViT.
 }
  \label{fig:fig1}
\end{figure}

\label{sec:intro}
Transformer architecture has led to the revolutionizing of Natural Language Processing (NLP) filed and inspires the emergence of transformer-type works on learning word~\cite{nlpword} and character~\cite{nlpcharacter} level representations with self-attention mechanisms~\cite{attentionisallyouneed} for capturing dependency syntax~\cite{nlpsyntax} and grammatical~\cite{nlplinguistic} relationships.
This has also motivated the recent works of vision transformers for vision tasks by using multi-head self-attention and multi-layer perceptrons, which are shown able to perform well on ImageNet classification and various downstream tasks, such as object detection~\cite{detr,rethinkingdetection,deformable}, semantic segmentation~\cite{segtrans,segformer,rethinktransseg}, etc.


Different from convolutional neural networks~\cite{hintoncnn,lecuncnn} which focus on local receptive fields, transformer-based architecture~\cite{vit} utilizes patch-wise attention mechanism for full-patch information interactions with dynamic receptive fields~\cite{naseer2021intriguing}.
As a general vision transformer backbone, ViT is capable of global feature extraction by dense information aggregation and interactions among full patch tokens. Several previous methods~\cite{transfg,attentionisallyouneed} have interpreted how classification outputs are formed in vision transformer models. 
Nevertheless, these visualization schemes mainly focus on the analysis of attention mechanism in discriminative patch selection or feature representations visualization, but remains unclear on the actual scope of 1-to-N patch attention. The valid question then emerges as is there redundancy in global self-attention?
%
%
Recently, there are some variations~\cite{regionvit,cswin,swin,dynamicvit} of ViT with heuristic configurations demonstrated the restricted attention range/window/region for patches can reduce redundancy but without performance degradation. Thus, the analysis of information interactions among patches during global self-attention become increasingly important, since it would play a crucial role in specifying the boundary of efficient attention scope, precisely dropping the indiscriminative patches or patch connections, and eventually guiding the visual attention model design.

To this end, we seek to obtain a better understanding of vision transformer models, especially the information interactions between patches.
The problem of understanding vision transformer presents various challenges due to its inherent architecture complexity. Specifically, the input image embedding features are learned across multiple layers, and utilize self-attention mechanism, that expresses an independent image patch output embedding feature as a convex combination of all patches embedding features. 
Meanwhile, the self-attention leverages multiple attention heads that operate independently. 
 
In this work, we propose a novel explainable visualization approach to analyze and interpret patch-wise interactions via quantifying the reliability of patch-to-patch connections, as shown in \cref{fig:fig1}.
Specifically, we propose a method for quantifying the impact of patch-wise attentions, in which patch token often gather information from the its related high-impact patches. In this way, we can briefly highlight the boundary of interactive regions for each patch. To verify the effectiveness of our quantification, we propose a novel adaptive attention window design schema yielding to the interactive region boundary for each patch. The experimental results show that this schema can improve ViT performance with reducing a large number of attention operations outside the attention window. Meanwhile, we observe that some indiscriminative patches provide their contextual information to all other patches consistently. Therefore, we design a mining schema for dropping these indiscriminative patches. To further understand the patch interactions, we define the responsive field for patch, and make statistic analysis on it. 
We find that the responsive field for each patch presents semantic relevance. This further motivated us to propose a window-free transformer architecture (WinfT) by incorporating the supervisions of responsive field. Correspondingly, the stable and significant performance improvements from window-free transformer further demonstrate the effectiveness of our analysis. The main contributions of this work are summarized below:

\begin{itemize}
[itemsep=2pt,topsep=0pt,parsep=0pt]
\item We propose a novel explainable visualization schema to analyze and interpret the crucial attention interactions among patches for vision transformer. For verifying the rationality of our visualization schema, we apply it to guide the attention window design and results in indeed performance improvement with significantly reducing the computational complexity.

\item Based on the quantification of the impact of patch interaction, we figure out the existence of indiscriminative patches in images for ViT model training. Further, dropping these patches also benefit the ViT model. 

\item We define responsive field for providing a crucial understanding of vision transformer. The statistic analysis shows that both the size and tendency of the informative attention window for each patch are semantic oriented.

\item  Inspired by above observations and analysis, we propose a novel window-free transformer architecture with predictive and adaptive attention window to restrict the patch-wise interactions. The experimental results in ImageNet show that our window-free transformer can improve $2.56\%$ top-1 accuracy while reducing nearly $58.9\%$ of patch-wise attention operations in average across various ViT structures with different input resolutions.

\end{itemize}

\section{Related Work}

\noindent\textbf{Transformer-based vision models}\quad 
Transformer-based vision models aim to utilize the attention mechanism to learn global dependencies representations.
ViT~\cite{vit} is the first convolution-free transformer-based vision architecture, which applied attention to a sequence of fixed-size non-overlapping patches.
This architecture is beneficial for exploring global contextual information and achieves high performance on downstream tasks.
Various transformer-based vision models~\cite{regionvit,twins,cswin,swin,pyramidvit} present the attention mechanism through heuristic patch-wise interactions, in particular, focus on different structures to enhance the patch information interactions for effective attention computation.
Beyond that, DynamicViT~\cite{dynamicvit} propose a dynamic token sparsification to prune the tokens. 
These models highlight the importance of information interactions, typically through heuristic patch-wise interactions.
Different from the above approaches for patch-wise interactions, our approach aims to propose adaptive patch-wise interactions for attention mechanism.

\begin{figure*}[!hbt]
 \centering
  \includegraphics[width=\linewidth]{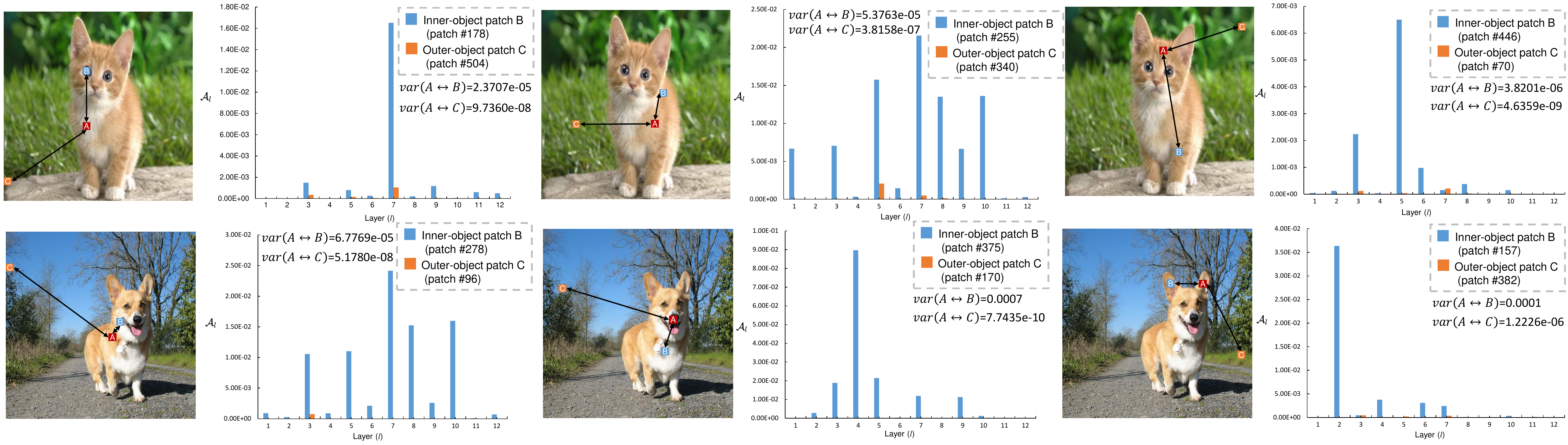}
 \caption{Illustration about quantitative analysis between the inner-object patch and outer-object patch. Zoom in for better visualization.}
  \label{fig:visualization_pair}
\end{figure*}

\noindent\textbf{Explainability for Vision Transformer}\quad 
Given the key role of explainability in deep learning, several works have analyzed the gradients~\cite{gradientgradcam,gradientnotjustablackbox,gradientsmoothgrad,gradientfullgradient,gradientaxiomatic} and attribute propagation~\cite{attributionLRP,attributeExplainingnonlinear} in Convolutional Neural Network to generate an understanding of representations to explain specific assumptions. 
Recent transformer works mainly~\cite{Leveragingredundancy,UnderstandingandImprovingRobustness,explain_bert_A,explain_bert_is_attention} focus on analyzing and interpreting attention scores to understand why the model performs so well. 
Reuse Transformer~\cite{Leveragingredundancy} highlights the relevant relationship between the different layers and captures the similarity to reuse the attention score.
Elena Voita et al.~\cite{heavylifting} apply layer-wise relevance propagation to consider the different relevance of multihead attention block.
These works are focused on visually understanding individual attention scores, and provide an explanation to understanding attention mechanisms.
However, there are two limitations to understand vision transformer models: i) they do not highlight the relevance of patch-to-patch connection. ii) focus on the individual model output, or attention scores, does not directly interpret the patch-wise interactions well.
We aim to provide an explainable visualization schema for ViT, therefore it's possible to analyze and interpret the interactions among patches.

\begin{figure*}[!hbt]
 \centering
  \includegraphics[width=\linewidth]{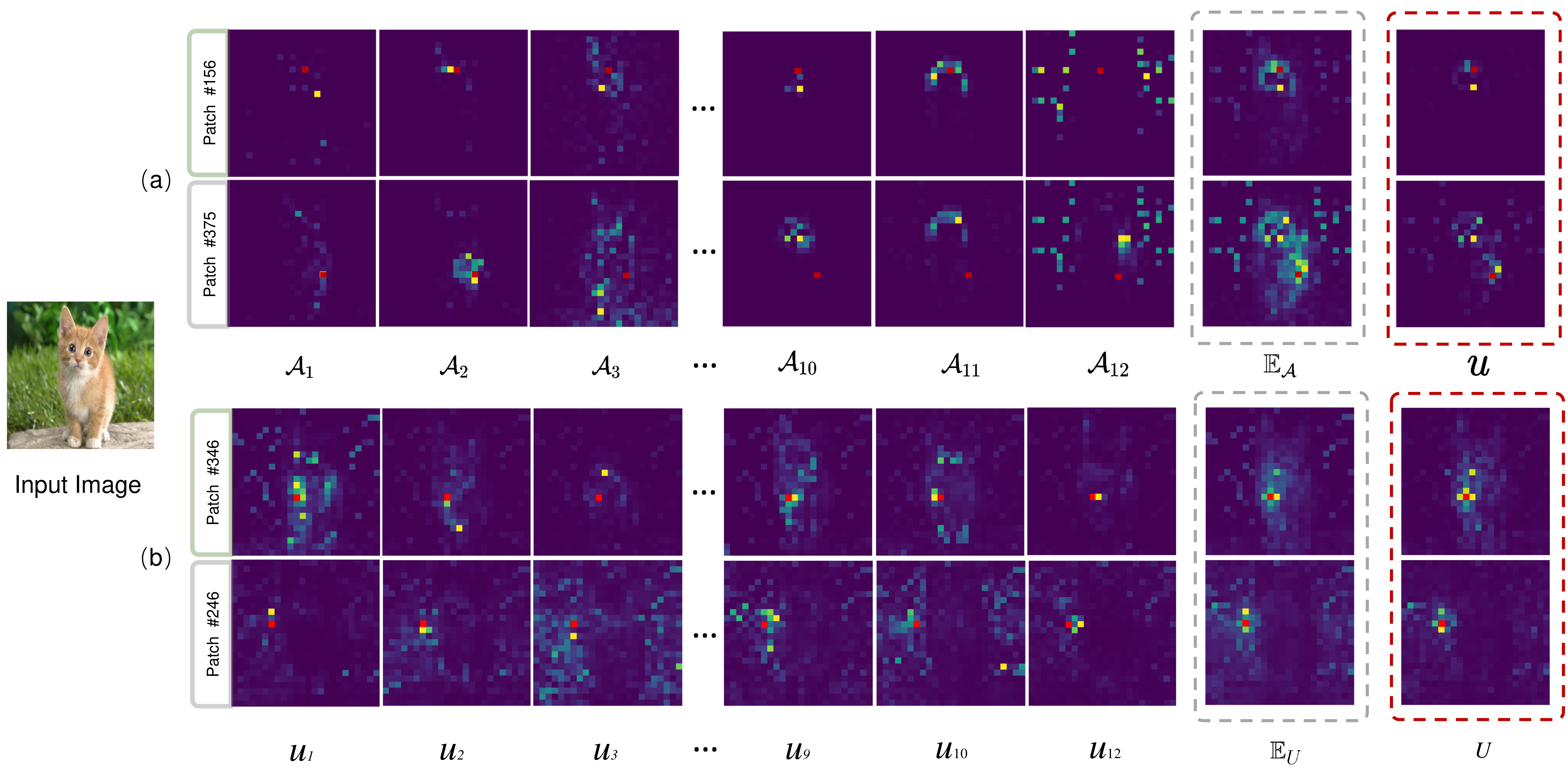}
 \caption{
   Visualization of attention score maps $\mathcal{A}$ and its uncertainty across different layers $\mathcal{U}$. 
   One $384\times 384$ resolution images are fed into the well-trained ViT-B/16~\cite{vit} with 12 heads, 12 layers and $16\times16$ patch size. We randomly sampled four patches (in \textcolor{red}{red} color) of the input image for visualization.
 }
  \label{fig:layer_unconcernty}
\end{figure*}

\section{Patch Interactions Visualization}
We study the patch interactions on global multihead self-attention of the original visual transformer~\cite{vit}, so we first present a brief background of ViT and describe our proposed visualization method for quantifying the magnitudes of patch-to-patch connections in vision transformers. 

\subsection{Preliminaries}

Given an input image of $H\times W$ resolution, Vision Transformer model (ViT) first splits it into a sequence of non-overlapping patches $\{\mathbf{x}_1,...,\mathbf{x}_N\}$ of fixed size $P$, and then transforms them into tokens by linear projection. 
For capturing the long-range dependencies among patches, the patch tokens are fed to stacked transformer encoders, each of which contains a multi-head self-attention (MSA) mechanism, which concatenates multiple scaled dot-product attention modules. Specifically, the scaled attention modules (SA) first linearly projects the patch tokens to a query matrix $Q$, key matrix $K$, and value matrix $V$, and then computes the attention weight matrix $\mathcal{A}$ according to the patch-wise similarity between the query and key matrices:
\begin{equation}
       \mathcal{A} = \text{SoftMax}({QK^T}/\sqrt{d}), \label{eq:attentionweight}\\
\end{equation}
where \(d\) is the channel dimension of the query or key. By computing the sum over $V$ weighted by row values $\mathbf{a}_p\in\mathbb{R}^{1\times N}$ in $\mathcal{A}$, information originating from different patch tokens get mixed for updating the representation of $p$-th patch token. 
Multihead self-attention (MSA) is an extension of SA in which concatenates $k$ self-attention operations. We denote the $k$-head attention weight matrix in MSA as $\mathcal{A}\in\mathbb{R}^{k\times N\times N}$. The interactions or contextual information exchanging among different patches mainly depends on the attention weight matrix $\mathcal{A}$. Thus, in this work, we focus on the theoretical and empirical analysis of attention weight matrix $\mathcal{A}$, rather than the representation of each patch or image region that widely used in previous visual feature analysis methods.

\subsection{Patch Interactions Quantification}
The transformer-based architectures are ideally capable of learning global contextual information by leveraging a fully self-attention mechanism among all patches.   
Extensive works~\cite{cswin,swin,dynamicvit,pyramidvit} have demonstrated the existence of redundant computations in the patch attention mechanism through fixed-scale window design, pruning, etc. However, these works usually focus on the spatial division or local region structuring on patches, while overlooked the reliability of the patch connections for self-attention mechanism. 
Meanwhile, through an analysis of similarity of attention scores by different layers and heads, the ability of interaction representation is not consistent~\cite{provearesixteenheads,provefixedencoder,dovisiontransformer,heavylifting}. We believe that reliability measure of patch connections by both structure and features would lead to a better understanding of the interactions among patches and further guide the attention mechanism design for ViT.

Many methods were suggested for generating a heatmap that indicates discriminative regions, given an input image and a CNN or transformer model. However, there are not many studies that explore the effectiveness of connections across image patches. Here, we start from visualization and statistic analysis on patch-wise interactions in ViT. As shown in \cref{fig:visualization_pair}, we randomly sampled target patch $A$ (in \textcolor{red}{red} color) inner object for analysis the patch interactions between the inner-object patch $B$ (in \textcolor{blue}{blue} color) and outer-object patch $C$ (in \textcolor[RGB]{237,125,49}{orange} color). Similar to the observation of ``ViT model’s highly dynamic receptive field'' that mentioned in previous work~\cite{naseer2021intriguing}, we can find that, although the irrelevance patch pair (inner-outer object patch pair) consistently has low response, the patch-to-patch attention score between relevance pair varies a lot across different layer, as the statistical results shown in \cref{fig:visualization_pair}. More intuitive examples can be found in \cref{fig:layer_unconcernty}(a), that the high attention areas cross all layers ($\mathcal{A}_{i}$ denotes the attention weight matrix in the $i$-th Transformer block) present an obvious uncertain of ``tight'' or ``loose''. This would be the potential cause of the flexible and dynamic receptive field of ViTs. Owing to the global information propagation among patch tokens in stacked multihead self-attention layers, a patch token can intensely gather representations from other patch tokens of high relevance with it. This results in unstable attention score and high uncertainty among relevance patches across layers. 

Thus, for a standard ViT model contains of $l$ multihead self-attention layer, all above observations and analysis motivated us to measure the impact of patch interactions by estimating the uncertainty of attention score among them across all Transformer blocks:
\begin{equation}
\mathcal{U}=\frac{1}{l}\sum_{i=1}^{l}\left(\mathcal{A}_{i}-\mathbb{E}_{\mathcal{A}}\right)^2.
\end{equation}
Thanks to the uniform head size setting of ViT model, we can directly compute $\mathbb{E}_{\mathcal{A}}\in\mathbb{R}^{k\times N\times N}$ as the mean of attention scores across all $k$-head self-attention layers. $\mathbb{E}_{\mathcal{A}}$ can be regarded as a $k$-channel score map depicting the uncertainty of interactions among patches. As we show in Fig.\ref{fig:layer_unconcernty} (a), although most of high uncertainty region are gathered around the corresponding patch, different channel of score map results in various range of uncertainty. Moreover, averaging the score map $\mathcal{U}$ cross channel (denoted as $\mathbb{E}_{U}$) results in signal attenuation of border areas. Thus, we apply another uncertainty estimation on $\mathcal{U}$ to quantify the various of uncertainty cross $k$ channel, and highlight the boundary of interactive regions for each patch as
\begin{equation}
U=\frac{1}{k}\sum_{i=1}^{k}\left(\mathcal{U}_{i}-\mathbb{E}_{U}\right)^2,
\end{equation}
where $\mathcal{U}_{i}$ is the $N\times N$ score map extracted from the $i$-th channel of $\mathcal{U}$. We visualize the row values $\mathbf{u}_p$ in $U$ for the given image patch $\mathbf{x}_p$ in \cref{fig:layer_unconcernty} (b). It can be found that most interactive regions of the current patch is in the surrounding area. There are also some connections of distant patches, which are usually relevant to background around the main object in image. These visualizations are also conform to the general knowledge that locality information play as a critical role for object recognition~\cite{twins,li2021localvit,swin}. Inspired by these related works, we also proposed two $U$ guided attention window design methods for ViT in \cref{sec:adaptive_window} and \cref{sec:windowfree_transformer} respectively. In special, we restricted the interaction range for each patch during 1-to-$N$ attention operations, by ignoring the patches outside the boundary of interactive regions in $U$.  The experimental results show that $U$ can well guide the ViT model training to focus on essential attention operation and lead to performance improvement while significantly decreasing the computational complexity. More details can be found in the sections below.

\section{Patch Interactions Analysis}\label{sec:patchwise_analysis}
Based on our quantification of patch-wise interactions, we figure out the existence of potentially indiscriminative patches for ViT model (\cref{sec:indis_patch}), and proposed an adaptive attention window design method (\cref{sec:adaptive_window}) for further analyzing the redundancy of global attention mechanism and the responsive field of each patch in ViT model (\cref{sec:responsive_field}).


\subsection{Adaptive attention window design}\label{sec:adaptive_window}
Attention mechanism is one of the core computations of the transformer-based model, requiring expensive quadratic calculations. Fixed-scale window/region design~\cite{regionvit,cswin,swin} and dynamic pruning structures~\cite{dynamicvit} are approaches to address redundancy in attention computation.
However, these approaches are extremely restrictive in the scale of the information interactions that need to be predefined artificially. Therefore, we propose a novel adaptive attention window design schema guided by the quantification result of patch-wise interactions for attention computation.

\begin{figure}[t]
 \centering
  \includegraphics[width=\linewidth]{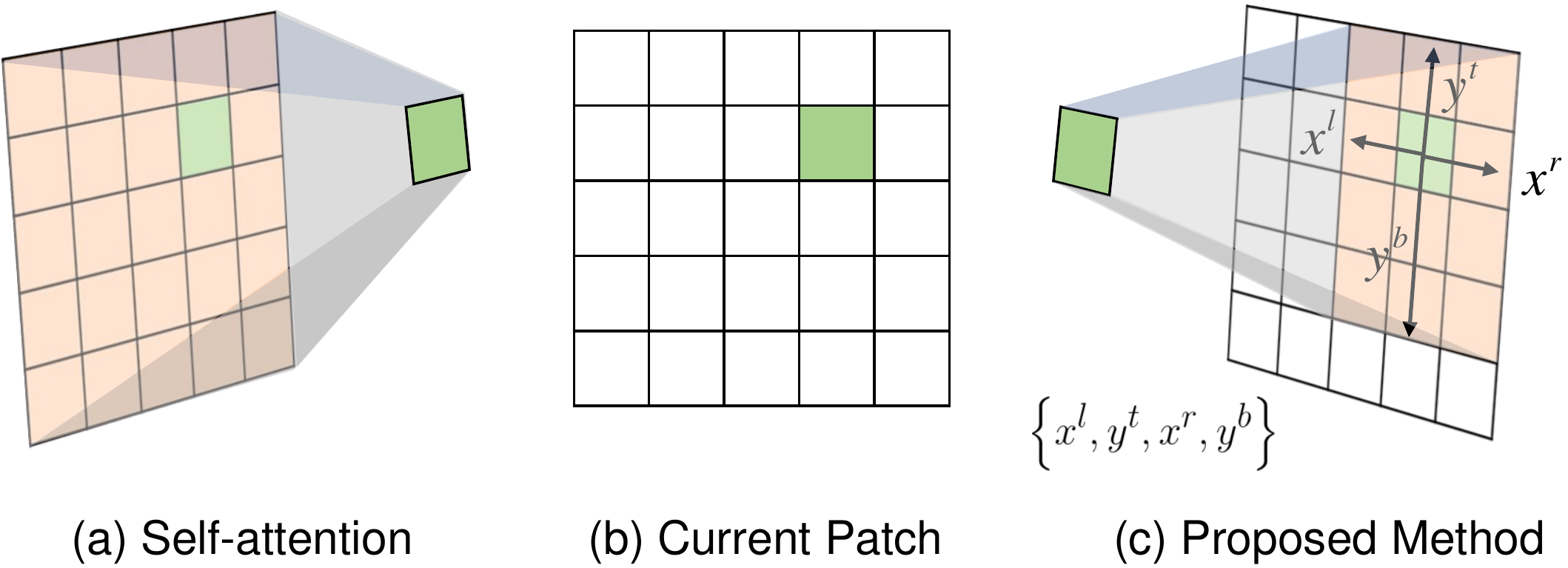}
 \caption{
   Illustration about global self-attention mechanism (a) and our proposed adaptive window design (c). Green patch represents the current patch for attention operation. Yellow region indicates the range of patch interactions for current patch.
 }
  \label{fig:wcompare}
\end{figure}

Considering $U$ can not only measure the informative of patch, but also highlight the boundary of interactive region for patch, we rank all values in $U$, and select the top $T$ elements with high value to construct a subset $U'$. After that, given the $p$-th patch whose coordinate is $\left \langle x_p, y_p \right \rangle$ ($x_p = p\%\sqrt{N}$, $y_p=p / \sqrt{N}$), we can construct a window boundary candidate set $B_p$ for it:
\begin{equation}
    B_p=\{\left \langle x_p, y_p \right \rangle\}\cup \left\{\left \langle x_i, y_i \right \rangle : u_{p,i}\in U'\right\}.
\end{equation}
Subsequently, we select the maximum and minimum offset in the x-axis and y-axis in $B_p$ respectively to finalize the attention window for $x_p$, denoted as $\{x^l_p, y^t_p, x^r_p, y^b_p\}$. Specifically, for the situation of $|B_p|=1$, $B_p=\{\left \langle x_p, y_p \right \rangle\}$ that there is no relevant patch hitting in $U'$, the global self-attention on $x_p$ degenerates into an identical operation. As a result, there are
\begin{equation}
    \mathcal{O} = \sum_{p=1}^{N}(x_p^r-x_p^l) \times (y_p^b-y_p^t)
\end{equation}
patch-wise interactions for each head in each self-attention layer. We denote $\mathbb{E}_{\mathcal{O}}$ as the averaged number of patch interactions per head over all images in dataset.

As shown in \cref{fig:wcompare}, the original self-attention mechanism leverages non-overlapping image patches and then build long-range interaction between all patches. Our proposed method provides adaptive attention window in terms of the effective interactions for each patch. Moreover, adaptive attention window design can reduce the redundant attention operation and decrease the complexity of ViT model.

\begin{table}[t]
\renewcommand{\arraystretch}{1.15}
\small
\centering
\begin{tabular}{l|c|c|c}
\hline\noalign{\smallskip}
Method & $\alpha$ & $\mathbb{E}_{\mathcal{O}}$ & Acc. (\%)\\
\noalign{\smallskip}
\hline
\noalign{\smallskip}
ViT-B/16\footnotemark & 1.0 & \makecell[r]{38,416}~ & 81.20\\\hline
\multirow{5}{*}{\shortstack{AWD-ViT-B/16 
}}~~ &0.50 &~38,334~ & 81.62 \\
&0.20 &10,283 & 81.25 \\
&0.10 &\makecell[r]{8,866}~ & 81.90 \\
&0.05 & \makecell[r]{6,801}~ & 80.60 \\
& ~0.025~ & \makecell[r]{4,971}~ & 78.28 \\
\hline
\end{tabular}
\caption{Top-1 accuracy of ViT on ImageNet by adapting adaptive window design (AWD) on various settings of $\alpha$. ViT-B/16 model are pretrained on ImageNet-21k and fine-tuned on ImageNet-1K at $224\!\times\!224$ resolution.}
\label{tab:topselect}

\end{table}
\footnotetext{The reported result in the official ViT implementation: \url{https://github.com/google-research/vision\_transformer}}


\textbf{Justification}\quad
Naturally, the quality of adaptive attention window design can directly reflect the rationality of our proposed uncertainty-aware quantification of patch-wise interactions. Thus we trained ViT models under various settings on ImageNet-1K dataset~\cite{imagenet} for justifying effectiveness of $U$. First, we computed the patch interaction score map $U$ for each image based on the well-trained ViT model, and then get the attention window based on $B_p$ for each patch. After that, we incorporate this priori attention window range into all self-attention operations for re-training (not finetuning) ViT model.
We selected $T=\alpha N^2$ with $\alpha=\{2.5\%, 5\%, 10\%, 20\%, 50\%\}$ to generate window at different scales. The experimental results are shown in~\cref{tab:topselect}. It can be found that best Top-1 accuracy for adaptive attention window design is achieved when $\alpha$ is set to $10\%$. 
In this case, we only use the nearly 23\% of patch connections for global self-attention, with the improvement of $0.7\%$ from the baseline $(81.20\%)$. 

The experimental results demonstrate that the responsive filed for each patch is unique and data-dependent. Without the full-patch global attention, there is no decrease in the performance of the model.
Based on the observation and analysis, we further prove that these approaches~\cite{regionvit,swin,dynamicvit} of region/window/local are designed in a reasonable way. Meanwhile, it also validated the rationality of our approach for patch-wise interaction quantification.

\begin{figure}[!t]
 \centering
   \includegraphics[width=1\linewidth]{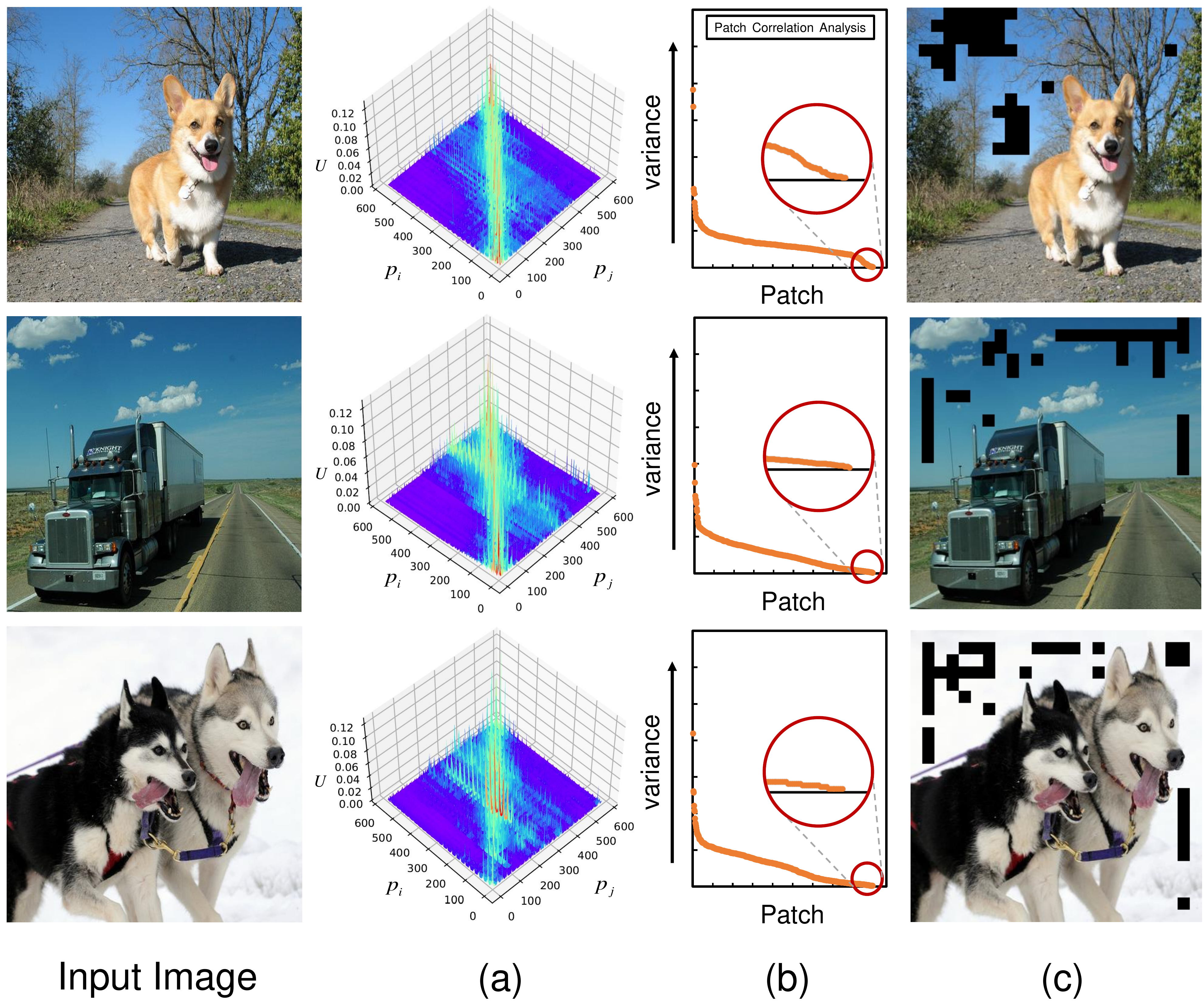}
 \caption{
  Illustration about analysis for the existence  of  potentially indiscriminative patches (Black) for ViT model.}
  \label{fig:biasanalysis}
\end{figure}

\subsection{Indiscriminative patch}\label{sec:indis_patch}

Noting that $U$ is not a symmetrical matrix. The raw values $\mathbf{u}_{p}=\{u_{p,1},...,u_{p,N}\}$ in $U$ measure the relevance of all $N$ patch tokens to the $p$-th patch token, while the column values $\mathbf{u}'_{p}=\{u_{1,p},...,u_{N,p}\}$ in $U$ reflects how informative $x_p$ is. Here we visualize the $N\times N$ score map of $U$ for a given image in 
~\cref{fig:biasanalysis} (a), and observe some anomalous patches with constant high column values of $U$ (in the red box). It means that these patches indiscriminately provide their information to all other patches from background to the main object in image. For a more intuitive explanation, we compute the variance of column values $\mathbf{u}'$ for each patch and sort the results in \cref{fig:biasanalysis} (b). We 
define the patch with low variance in $\mathbf{u}'$ as indiscriminative patches and visualize the indiscriminative patches for three different images (\cref{fig:biasanalysis} (c)). 
Obviously, these indiscriminative patches mainly located at empty information area in background of the key objects in image. More visualization results can be found in the Appendix.

Since the indiscriminative patches are data-dependent, but provide their contextual information to all other patch tokens consistently, they can be also regarded as the data-dependent bias for ViT model training.

\begin{figure}[t]
 \centering
    \includegraphics[width=1\linewidth]{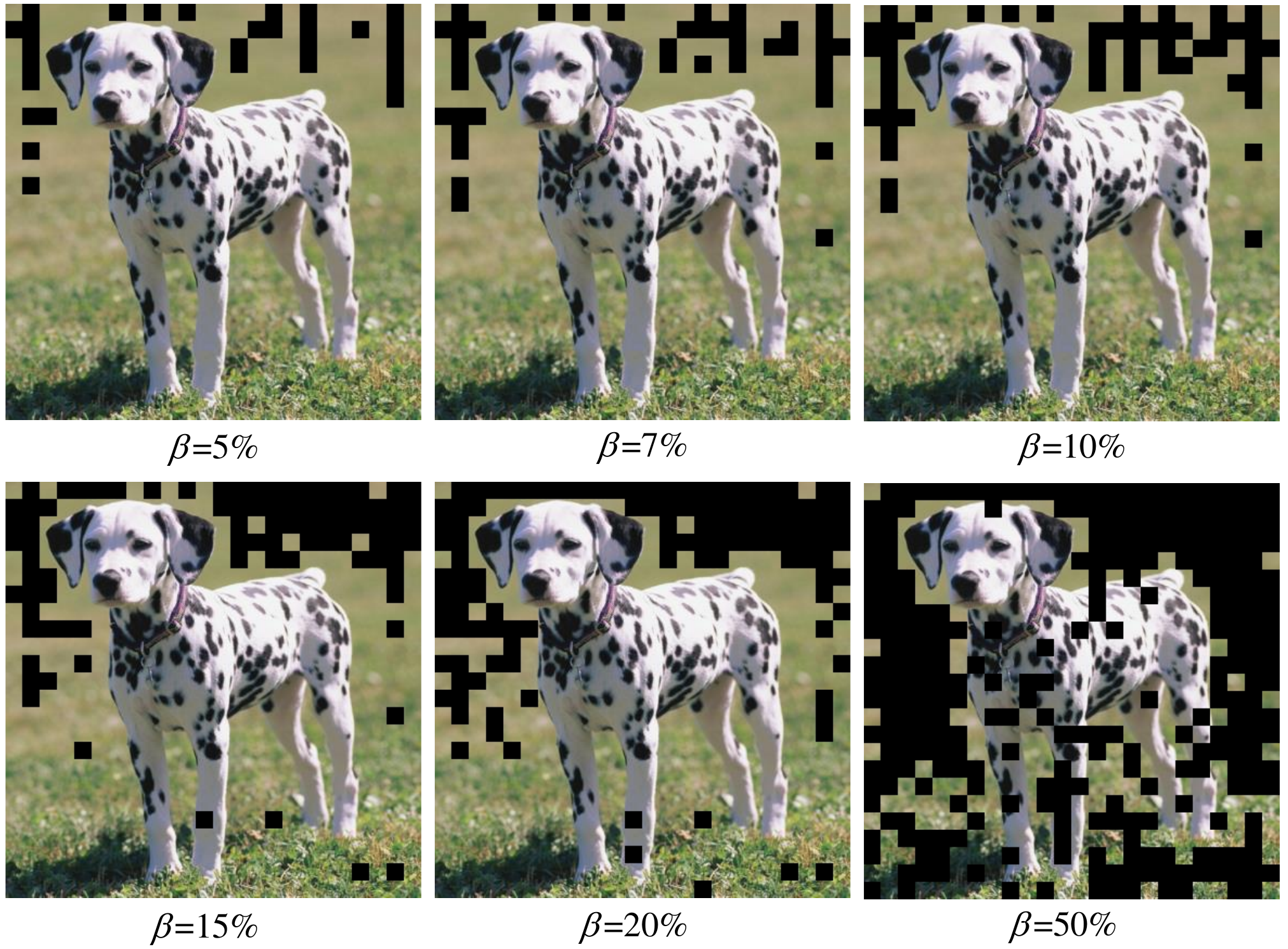}

 \caption{
  Illustration of dropping indiscriminative patches (Black) on various $\beta$ rates for ViT model.}
  \label{fig:dropbias}
\end{figure}

\textbf{Justification}\quad To understand how the indiscriminative patches impact the ViT model training, we retrained the adaptive attention window designed ViT models of $\alpha=0.10$ by erasing the indiscriminative patches. In special, we ranked the variance of $\mathbf{u}'$ for each patch, and generate a mask matrix $M$, where the $\beta N$ patches which has the lowest variance values are masked as 0, while other patches are masked as 1. We multiply $M$ for all patch tokens across all layers in ViT model during training and inference. The experimental results of ViT on ImageNet can be found in \cref{tab:bias}. Moreover, \cref{fig:dropbias} provides visualization about the influence of $\beta$ rates for dropping indiscriminative patches.

\begin{table}
\renewcommand{\arraystretch}{1.15}
\small
\centering
\begin{tabular}{l|c|c|c|c}
\hline\noalign{\smallskip}
Method & $\alpha$ &$\beta$ & $\mathbb{E}_{\mathcal{O}}$ & Acc. (\%)\\
\noalign{\smallskip}
\hline
\noalign{\smallskip}
ViT-B/16 & 1.0 & 0 & 38,416 & 81.20 \\\hline
\multirow{2}{*}{AWD-ViT-B/16} & 0.10 & 0 & \makecell[r]{8,866} & 81.90 \\
& 0.05 & 0 & \makecell[r]{6,801} & 80.60 \\\hline
\multirow{3}{*}{\makecell[l]{AWD-ViT-B/16 \\\textbf{\textit{w/ DIP}}}}
& 0.10 & 0.1 & \makecell[r]{7,982} & 82.09 \\
& 0.10 & 0.2 & \makecell[r]{7,080} & 82.40\\
& 0.10 & 0.5 & \makecell[r]{4,420} & 80.62 \\
\hline
\end{tabular}
\caption{Top-1 accuracy of Vision Transformer on ImageNet validation set by adapting our proposed AWD and drop indiscriminative patches (DIP) on various settings of $\alpha$ and $\beta$.}
\label{tab:bias}
\end{table}

\begin{figure*}[!t]
 \centering
  \includegraphics[width=\linewidth]{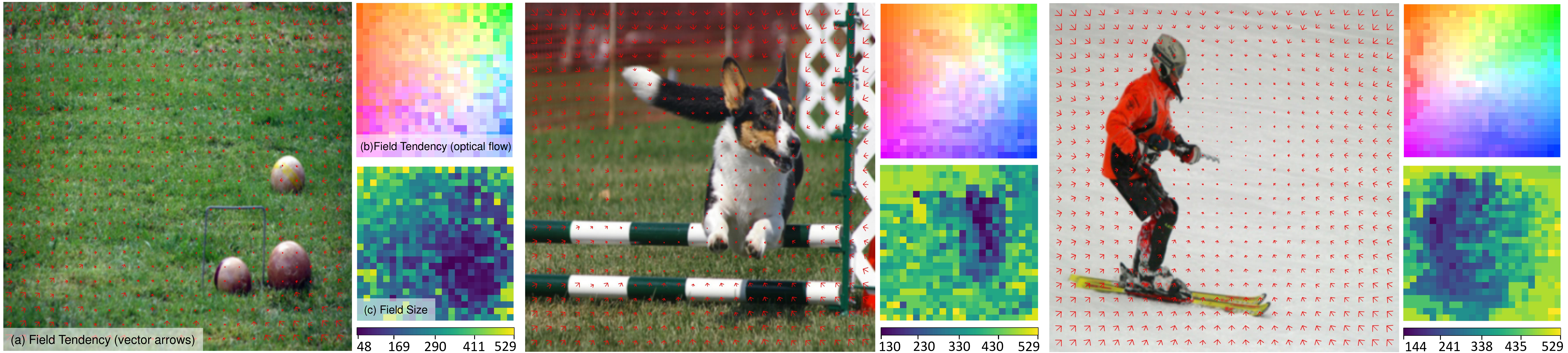}
 \caption{
    Examples of responsive field analysis from ViT model. We illustrate the responsive field tendency using optical flow tool~\cite{flow} and vector arrows. Meanwhile, we visualize the size of the responsive field for each patch using a heatmap.}
  \label{fig:Tendency} 
  \label{fig:Tendency_result}
\end{figure*}

It shows that dropping indiscriminative patches during ViT model training and inference results in the better final performance improvement (82.40 vs. 81.90). Even after masking out half of patches ($\beta=0.5$), the results of AWD-ViT-B/16 \textit{w/ DIP} still have comparable performance with the original ViT. Considering patches with low variance in $\mathbf{u}'$ fairly provide information to all patches during global attention operation, and they are also data-dependent (different image results in different indiscriminative patches distributions), we can treat these indiscriminative patches as the image-specific bias during model training. Such bias would mislead the ViT model to learn image identification rather than the general discriminative patterns. In general, for such indiscirminative patches in ViT, less is more.

\subsection{Responsive field analysis}\label{sec:responsive_field}

We define the adaptive attention window updated by dropping indiscriminative patches as the responsive field for each patch. Here we make statistic analysis on responsive field in the following two aspects:

\textbf{Field size}\quad Following the adaptive window design, given the $p$-th patch coordinate $\left \langle x_p, y_p \right \rangle$, its attention window offsets $\{x^l_p, y^t_p, x^r_p, y^b_p\}$, indiscriminative patch set $D$ of current image, the responsive field of $\mathbf{x}_p$ can be expressed as:
\begin{equation}
    S_p=\left\{\mathbf{x}_p\right\} \cup \left\{\mathbf{x}_i : \left \langle x_i, y_i \right\rangle \in B_p, \mathbf{x}_i \notin D \right\}.
\end{equation}
Here we visualize the size of responsive field $|S_p|$ for each patch in Fig.~\ref{fig:Tendency_result} (c). We observe that the distribution of patch's responsive field size is relevant to the semantic information of each patch, i.e. responsive field for patch of the main object usually tend to be smaller that the patch of background. A consequence is that smaller responsive fields are more focused on the local texture or structure learning, while big responsive field aims to learn the correlation between object and the background.

\begin{figure}[!t]
 \centering
  \includegraphics[width=\linewidth]{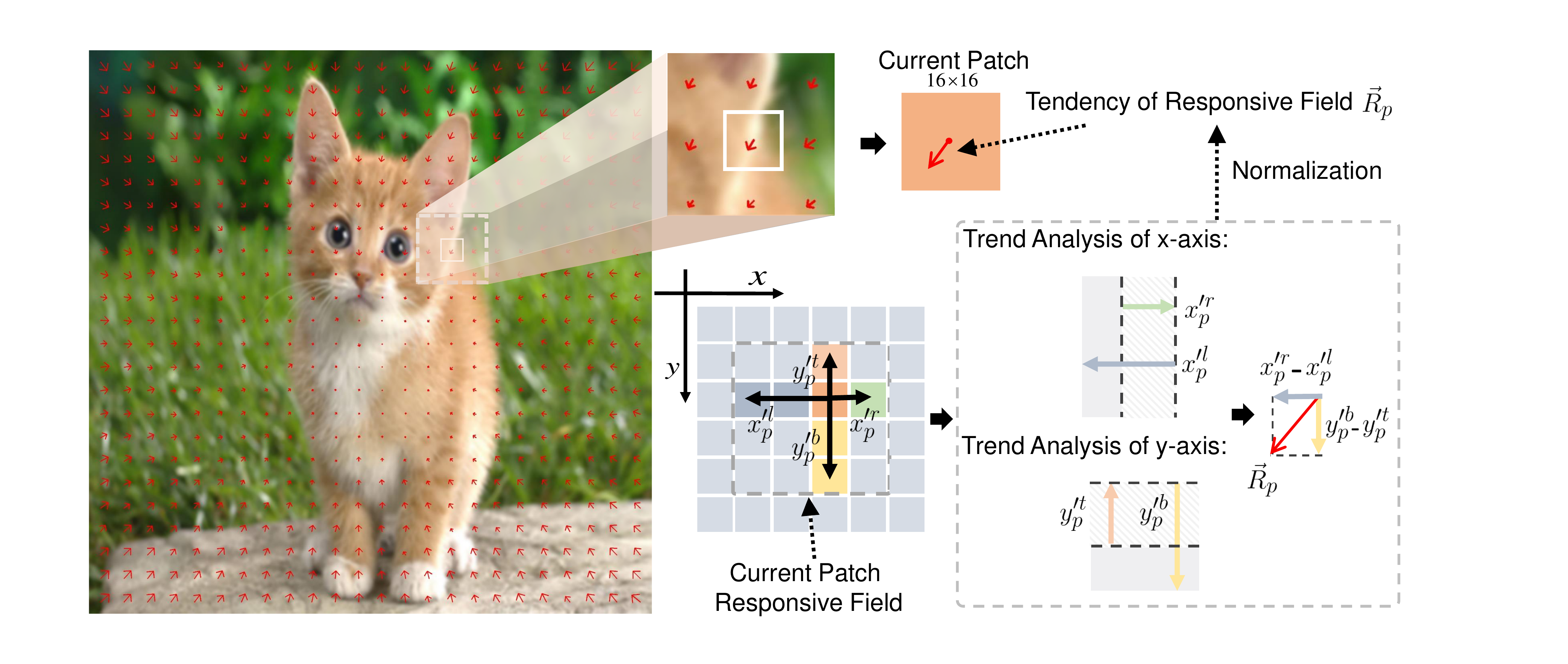}
 \caption{
  Illustration of the tendency and size analysis of responsive field for given patch.}
  \label{fig:Tendency_size}
\end{figure}

\textbf{Field tendency}\quad 
Meanwhile, the responsive field of each patch is constrained with four directions offsets in x-axis and y-axis. The patch-wise interactions can calculate the current patch tendency of responsive field as shown in Fig.~\ref{fig:Tendency_size}. Thus, we compute the $p$-th patch responsive field $\vec{R}$:
\begin{equation}
\vec{R}_p=( x'^r_p-x'^l_p, y'^b_p-y'^t_p),
\end{equation}
where ${x'^r_p, x'^l_p, y'^b_p, y'^t_p}$ are the maximum and minimum offset in the x-axis and y-axis of responsive field $S_p$ respectively. After that, we normalize the $\vec{R}_p$ to represent the tendency of $p$-th patch responsive field, and then visualize the tendency of each patch responsive field. As shown in Fig.~\ref{fig:Tendency_result}, similar to the visualization of field size, the 
tendency of responsive field are also semantically relevant. Overall, the directional field of $S_p$ is object-centric. 

\section{Window-free Transformer}\label{sec:windowfree_transformer}
The patch-wise interactions analysis provides a novel complementary view to understanding the vision transformer model. Based on our observation and analysis, we propose a simple yet transformer architecture by incorporating the supervision of responsive fields during training.
Meanwhile, this architecture can further validate the effectiveness of our observations.

\subsection{Window-free multihead attention}
Multihead attention mechanism is the core of transformer architecture. Although much work has focused on the attention scores and layers to explore the representational ability in transformer, understanding which interactions are most effective that may influence effectiveness is critical to achieve improvement.
Therefore, following our understanding and visual analysis, we design a data-driven multihead attention mechanism by incorporating the supervision of responsive fields during training.

\begin{figure}[!t]
  \includegraphics[width=\linewidth]{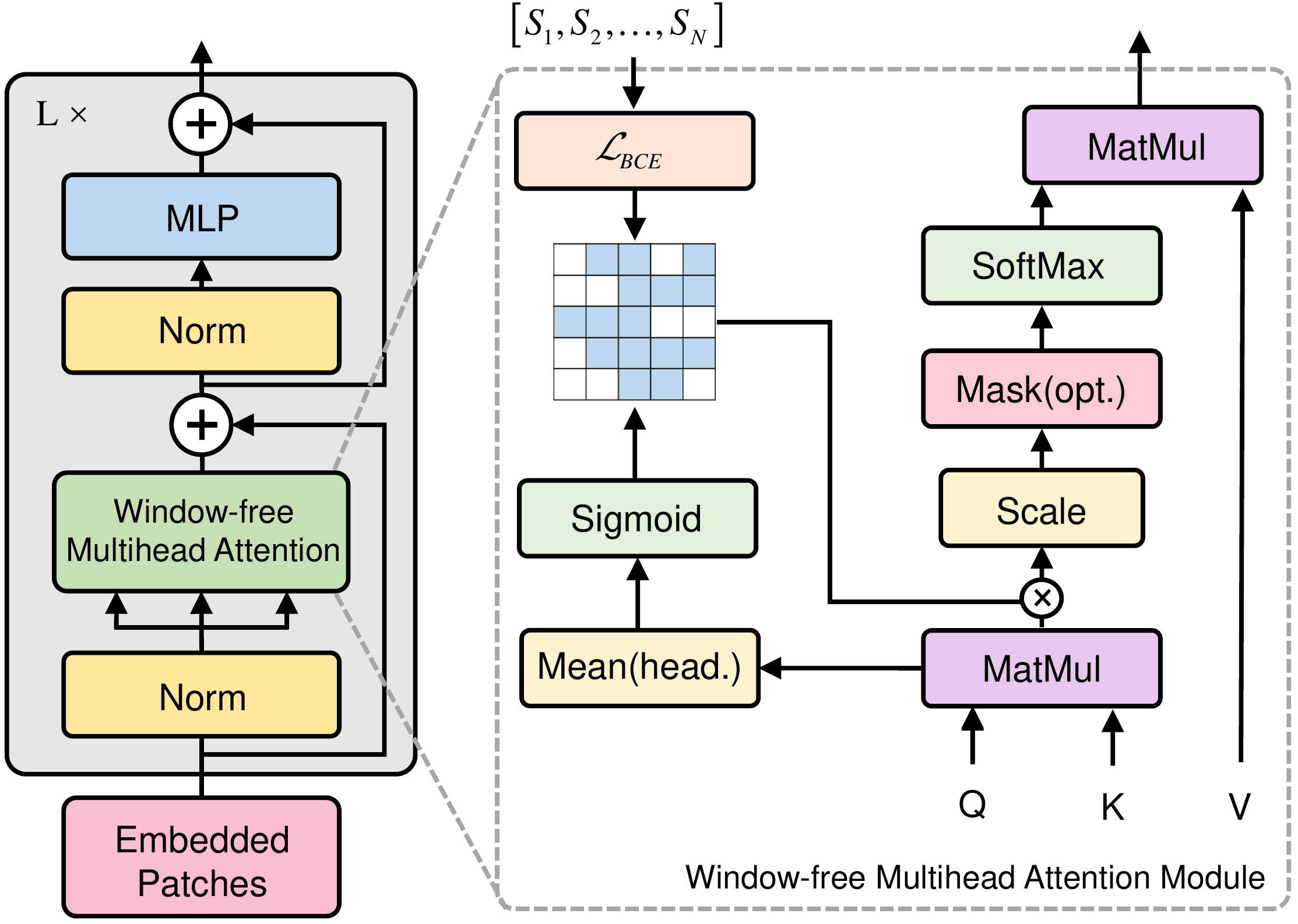}
 \caption{
   Illustration of a Transformer block with our Window-free Multihead Attention module. 
   This attention mechanism is trained with external supervisions of the responsive field for presenting an adaptive patch-wise interaction window. No external knowledge need during inference.}
  \label{fig:winfmodule}
\end{figure}

As shown in~\cref{fig:winfmodule}, the window-free module first linearly projects the patch tokens to $Q$, $K$ and $V$ as inputs.
we compute the dot products of the $Q$ with all $K$. After that, we apply an average computation and a sigmoid function to obtain the weights $w'$ for window design on the values $QK^T$. Then, we can generate a binary mask $W$ through the weights $w'$ for restricting the patch-wise interactions.
%
%
Specifically, we formulate this process as follows:
\begin{align} 
\begin{matrix}
w'=\text{Sigmoid}(\frac{1}{k}\sum_{}^{k}(QK^T)),\\
\\
W_{i,j}=\begin{cases}
1, & w'_{i,j}> \overline{{w'_{i,:}}} \\
0, & w'_{i,j}\leqslant \overline{{w'_{i,:}}},
\end{cases}
\end{matrix}
  \label{eq:attention}
\end{align}
where $i$ and $j$ are the horizontal and vertical, respectively. ($1\leqslant i,j\leqslant N$). $\overline{{w'_{i,:}}}$ represents as a dynamic threshold to get binary mask, which is the average of all values $w'_{i,1:N}$ from the $i$-th patch in the weights $w'$.

Therefore, the window-free multihead attention module computes the element-wise products between the $QK^T$ and $W$, scales to stabilize training, and then softmax normalizes the result. The final attention results are obtained by computing dot production of value matrix $V$ with masked attention score matrix: 
\begin{equation}
\mathcal{A}'=\text{SoftMax}(\frac{QK^T}{\sqrt{d}}* W),
  \label{eq:attention}
\end{equation}
where $d$ is embedding dimension of $Q$ and $K$.


\subsection{Window ground-truth and loss}
\textbf{Window ground-truth}\quad 
We use the patch-wise interactions analysis tools to evaluate all training dataset images.
The outputs of image \(I\) patch-wise interactions window groundtruth \(w_{gt}\) can be written as:
\begin{equation} 
w_{gt}(I) = \Theta_{(0,1)} ( \left [ S_{1}, S_{2}, \cdots, S_{N} \right ]),  
  \label{eq:wi}
\end{equation}
where $\Theta_{(0,1)}$ represents the conversion of each patch's responsive field $S_{p}$ into a binary mask.
Note that we convert all window offsets to binary masks for more efficient attention computation.

\textbf{Loss function}\quad
To give a clear hint, we introduce the patch-wise interactions mask to guide the adaptive window design via adding a binary cross-entropy (BCE) loss between the window-free module output $w'$ and corresponding interactions window groundtruth binary mask $w_{gt}$ :
\begin{align} 
\mathcal{L}_{BCE}=-\frac{1}{N^2}\sum (w_{gt} log(w')+(1-w_{gt})log(1-w')).
  \label{eq:lossmse}
\end{align}
$\mathcal{L}_{BCE}$ provides a learnable adaptive window design representation of patch-wise interactions. By doing so, our window-free transformer architecture adaptively captures the responsive field.

We adopt cross-entropy loss \(\mathcal{L}_{CE}\) as classification loss.
Therefore, our model is trained with the sum of  \(\mathcal{L}_{CE}\) and \(\mathcal{L}_{BCE}\) together which can be formulated as:
\begin{align} 
\mathcal{L}_{total} = \lambda_{1} \mathcal{L}_{CE} + \lambda_{2} \mathcal{L}_{BCE},
  \label{eq:lossall}
\end{align}
where the \(\lambda_{1}\) and \(\lambda_{2}\) are hyper-parameters, which are respectively set to 1 and 1 in our experiments.

\subsection{Experimental Results}
\textbf{Results on ImageNet}\quad
Following the settings in ViT~\cite{vit}, we adopt ViT-B as our backbone, which contains 12 transformer layers in total and pretrained on ImageNet-21K with $16^2$ or  $32^2$ patch size. The batch size is set to 512. And we train all models using a mini-batch Stochastic Gradient Descent optimizer with the momentum of 0.9. The learning rate is initialized as 0.01 for ImageNet-1K. We then apply cosine annealing as the scheduler for the optimizer.

\begin{table}
\renewcommand{\arraystretch}{1.15}
\centering
\small
\begin{tabular}{l|c|c|c}
\hline\noalign{\smallskip}
Method  & image size &  Acc. (\%) &  $\mathbb{E}_{\mathcal{O}}$\\
\noalign{\smallskip}
\hline
\noalign{\smallskip}
 ViT-B/32 & 224$^2$ &  74.46 & \makecell[r]{2,401}\\
 ViT-B/32 & 384$^2$ &  81.28 & \makecell[r]{20,736} \\
 ViT-B/16 & 224$^2$ &  81.20 &\makecell[r]{38,416}\\
  ViT-B/16 & 384$^2$ &  83.61 &\makecell[r]{331,776}\\
\hline
 WinfT-B/32 & 224$^2$ & 78.74  &  \makecell[r]{906}\\
 WinfT-B/32 & 384$^2$ & 84.33  & \makecell[r]{8,638}\\
 WinfT-B/16 & 224$^2$ & 83.11 & \makecell[r]{16,938}\\
 WinfT-B/16 & 384$^2$ & 84.62  & \makecell[r]{136,327}\\
 \hline
\end{tabular}
\caption{Top-1 accuracy comparison with ViT methods on ImageNet. The $\mathbb{E}_{\mathcal{O}}$ of WinfT measures the sum of predicted binary patch-wise attention mask $W$ averaged over all images.}
\label{tab:window}
\end{table}

We notice in~\cref{tab:window} that adaptive window learning to restrict the patch-wise interactions consistently lead to better performance. Specifically, compared with different input image resolutions and patch sizes, the window-free transformer architecture can achieve better performance with strong correlation patch connections for interactions. An interesting point is that after fitting suitable attention window, the ViT model with $32\times 32$ patch size can achieve similar performance with the settings of $16\times 16$ patch size (84.33\% vs. 84.62\%). Bigger patch size with less patch amount results in substantially reducing computation complexity for self-attention operation. It benefits the practical application of vision transformer models. In general, these experimental results validate the effectiveness of the supervision of responsive field and our proposed quantification method for the impacts of patch-wise interactions.

\textbf{Results of transfer learning}\quad
To further verify the effectiveness of our proposed visualization analysis and window design method, we conduct a comprehensive study of fine-grained classification. Note that fine-grained classification aims at classifying the sub-classes to find subtle differences in similar classes, and the model needs to focus more on discriminative feature learning.

\begin{table}
\renewcommand{\arraystretch}{1.15}
\centering
\small
\begin{tabular}{l|c|c}
\hline\noalign{\smallskip}
Method & Acc. (\%) & $\mathbb{E}_{\mathcal{O}}$  \\ 
\noalign{\smallskip}
\hline
\noalign{\smallskip}
ViT-B/16 & 90.30 & \makecell[r]{614,656} \\ \hline
WinfT-B/16 & 90.58 & \makecell[r]{207,532}\\
\hline
\end{tabular}
\caption{Performance comparison on CUB dataset. All models are trained and evaluated at $448\!\times\!448$ resolution.} 
\label{tab:cub}
\end{table}




We show the experimental results of transfer learning on fine-grained benchmarks in~\cref{tab:cub}. 
Our WinfT achieves 0.28\% improvement on Top-1 accuracy with reducing 66.24\% patch-wise interactions compared with the results of the original ViT. It further demonstrated the generalization of our proposed patch interaction analysis method.

\section{Limitations}
The visual analysis and understanding provide a more interpretive understanding of the patch-wise interactions and further guide the design of effective the transformer architectures, but there still exist some limitations.
%
Inevitably, we need to retrain the model through the adaptive window design to restrict the patch-wise interactions.
Notably, we need to state that the proposed method is aimed at verifying the reasonableness of the analysis of the patch-wise interactions instead of proposing a state-of-the-art transformer architecture. The computation reduction in $\mathbb{E}_{\mathcal{O}}$ reflects the potentiality of higher efficient ViT models, but WinfT still needs to compute the attention mask. Our window-free transformer model can be treated as distilling a model using the less attention operations. More importantly, we hope that our analytical methodology can provide some new insights for future transformer-based model design. 

\section{Conclusions}
In this paper, we proposed a novel explainable visualization schema to analyze and interpret the patch-wise interactions for vision transformer.
Concretely, we first investigate the interaction between patches and then propose the uncertainty-aware quantification schema for measuring the impact of patch interactions. Based on the quantification results, we make a series of experimental verification and statistic analysis on the responsive field of patch. Motivated by our observations on responsive field, we propose a window-free transformer architecture by adaptively restricting the patch interactions. Experimental results demonstrated both effectiveness and efficient of our architecture.

{\small
\balance
\bibliographystyle{ieee_fullname}
\bibliography{egbib}
}

\newpage
\onecolumn
\section*{A.Supplementary Material}
This supplementary material provides more visualization analysis samples to verify our visualization and understanding of patch interactions in vision transformer. First, we present details of our experimental setup. Then, the responsive field's tendency and size distributions are also visualized. Together, we provide more extended visualization results by analyzing the patch-wise interactions to prove the efficacy of our proposed methods further. After that, we compared the supervision of the responsive field from ViT and the predicted attention mask in WinfT. 
\section*{1.Experimental Details}
\begin{figure}[!h]
\centering
		\includegraphics[width=0.98\linewidth]{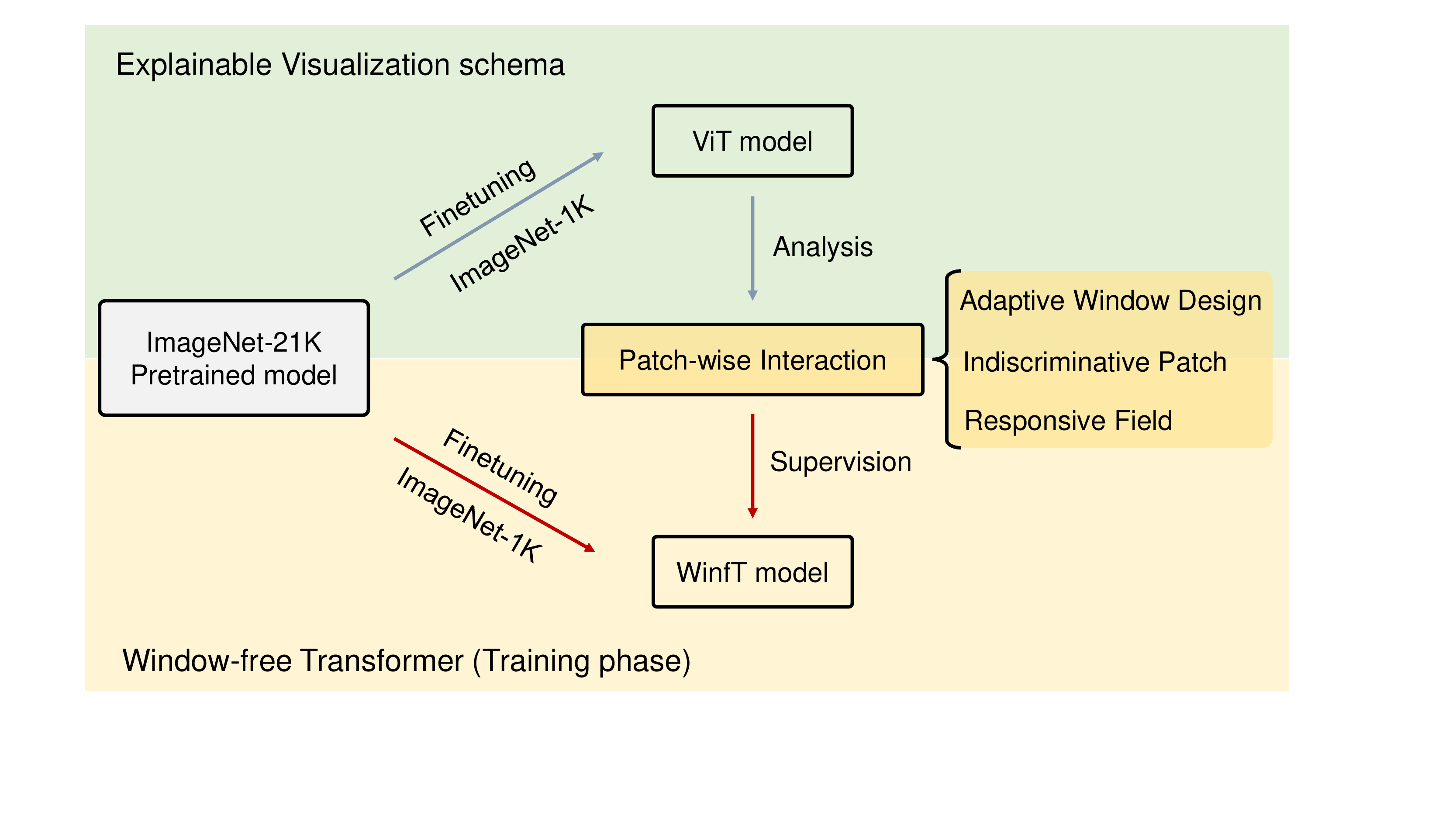}
    \caption{Schematic illustration of the experimental procedure. Notably, \textbf{the WinfT model do NOT need the supervision of responsive fields during inference}.}
    \label{fig:overall_pipeline}
\end{figure}

As shown in \cref{fig:overall_pipeline}, our explainable visualization schema aims to analyze and interpret the patch-wise interactions, and then extract valuable patch interaction supervisions for guiding WinfT model training. It is worth noting that the WinfT model does not require any supervision information during the inference phase. 

\textbf{Implementation on ImageNet} We use the intermediate weights from official ViT-B/16 and ViT-B/32 models pretrained on ImageNet-21K.
The training is conducted with 4 GPUs (gradient accumulation is applied owing to the limited GPU memory) with mini-batch size of 512 and initial learning rate of 0.01. We use SGD as the optimizer, the momentum of SGD is set as 0.9. The model is trained for 15 epochs. We use 500 warmup steps and adopt cosine annealing as the scheduler of the optimizer. All these settings stay the same for our re-implemented ViT and our WinfT.

\textbf{Implementation on CUB-200-2011} We finetuned the ViT and WinfT model on CUB dataset from the official ViT-B/16 model pretrained on ImageNet-21K. These models are trained for 20,000 steps with a batch size of 16 and initial learning rate of 0.03. Cosine annealing is adapted as the learning rate scheduler of optimizer. 

\section*{2.Visualization Analysis of Responsive Field}
\begin{figure*}[!h]
\centering
		\includegraphics[width=0.8\linewidth]{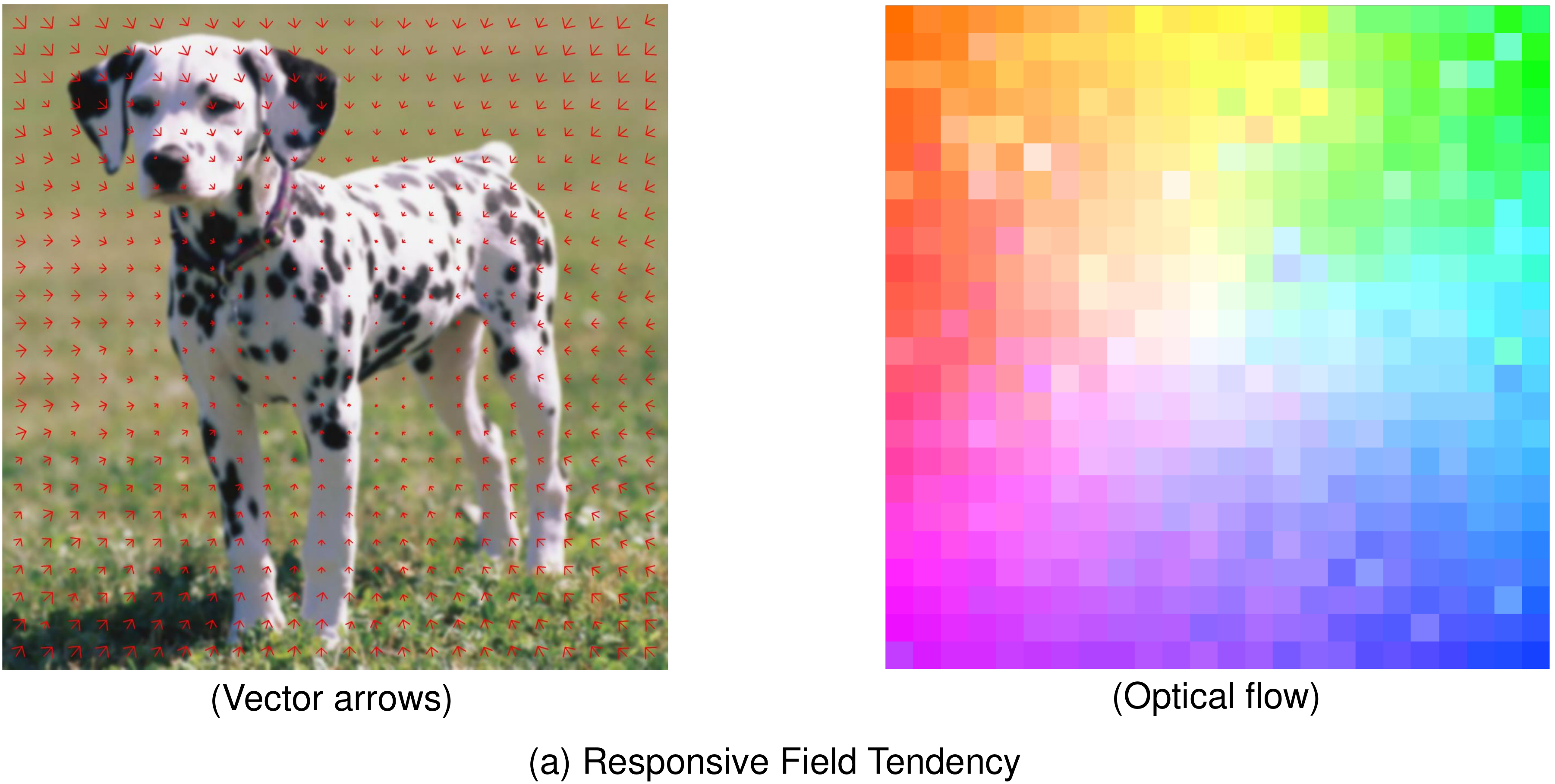}
	 \caption{The visualization interpretation of the responsive field tendency by using vector arrows and optical flow.}
	 \label{fig:tendency}
\end{figure*}

\begin{figure*}[!h]
\centering
		\includegraphics[width=0.9\linewidth]{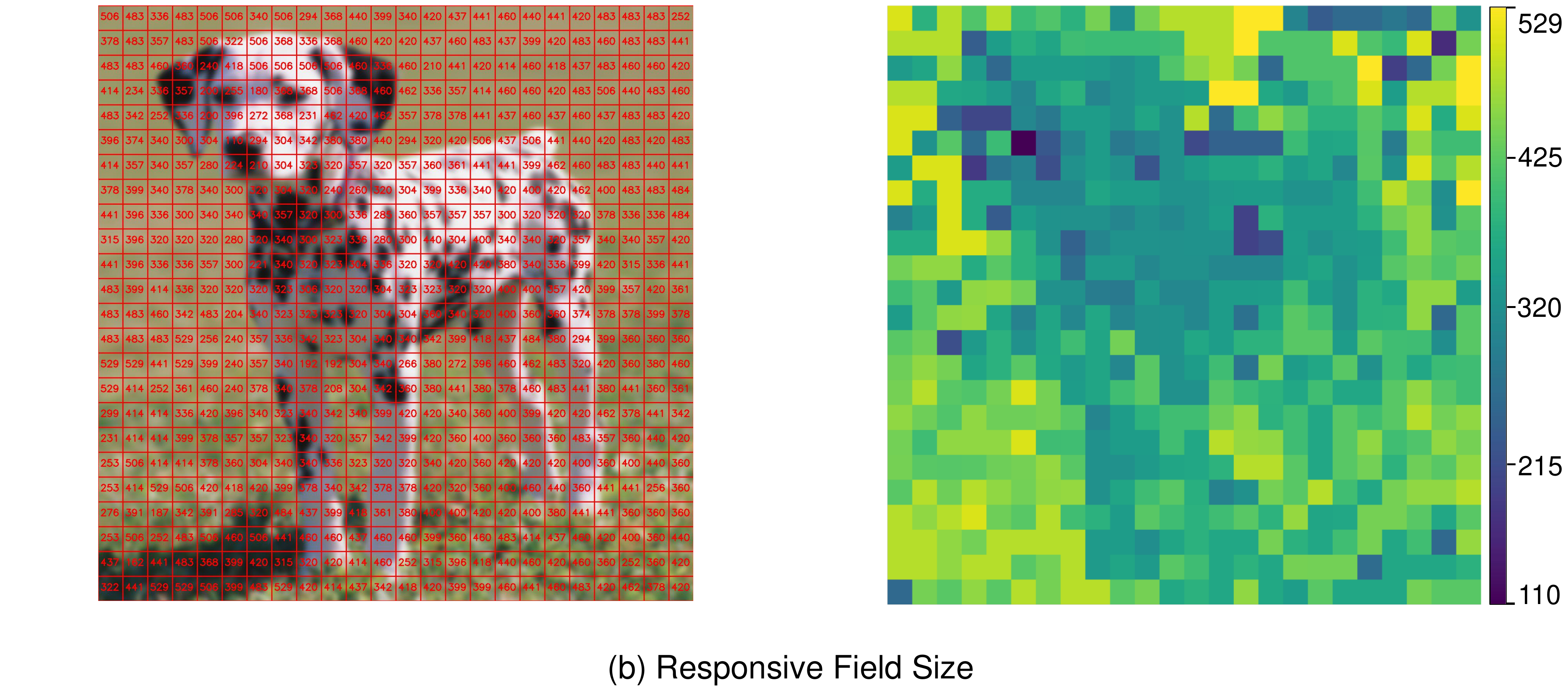}
	 \caption{The visualization interpretation of the size of responsive field.}
	\label{fig:size}
\end{figure*}

We adopt the adaptive window design schema and indiscriminative patches to define the responsive field (Sec. \textcolor{red}{4.3}), and further analyze and interpret the patch-wise interaction. Then, we utilize the statistic analysis of responsive field tendency and size for studying the patch interactions in ViT. \cref{fig:tendency} and \cref{fig:size} are the higher resolution versions of Fig. \textcolor{red}{6}. We visualize the tendency by directly drawing the direction vector of each patch's responsive field as a pixel in the optical flow.

\section*{3.Extended Visualization Results}
Combined, we visualize more samples in ImageNet dataset to verify the efficacy of our proposed explainable visualization schema below.

\begin{figure*}[!b]
\centering
	\vspace{-0.5cm}
		\includegraphics[width=\linewidth]{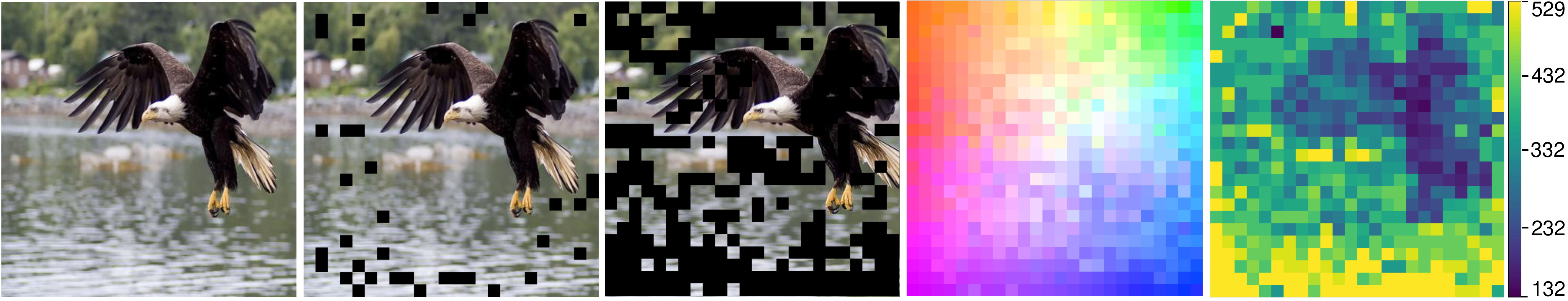}
	\label{sm_vis_area5}
\end{figure*}
%
%
\begin{figure*}[!p]
\centering
		\includegraphics[width=\linewidth]{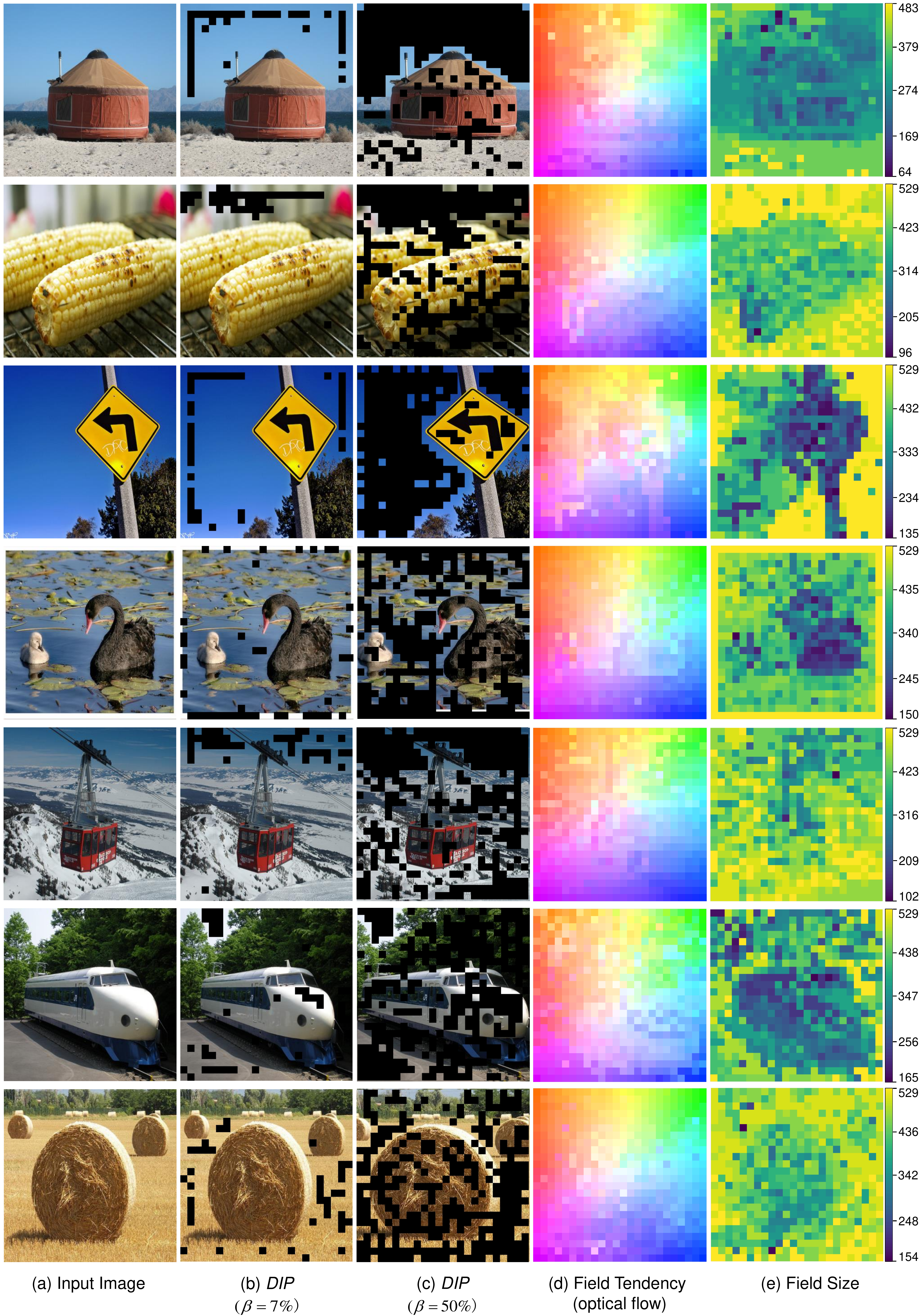}
	\label{scannet_vis}
\end{figure*}

\begin{figure*}[!b]
\centering
	\vspace{-0.5cm}
		\includegraphics[width=\linewidth]{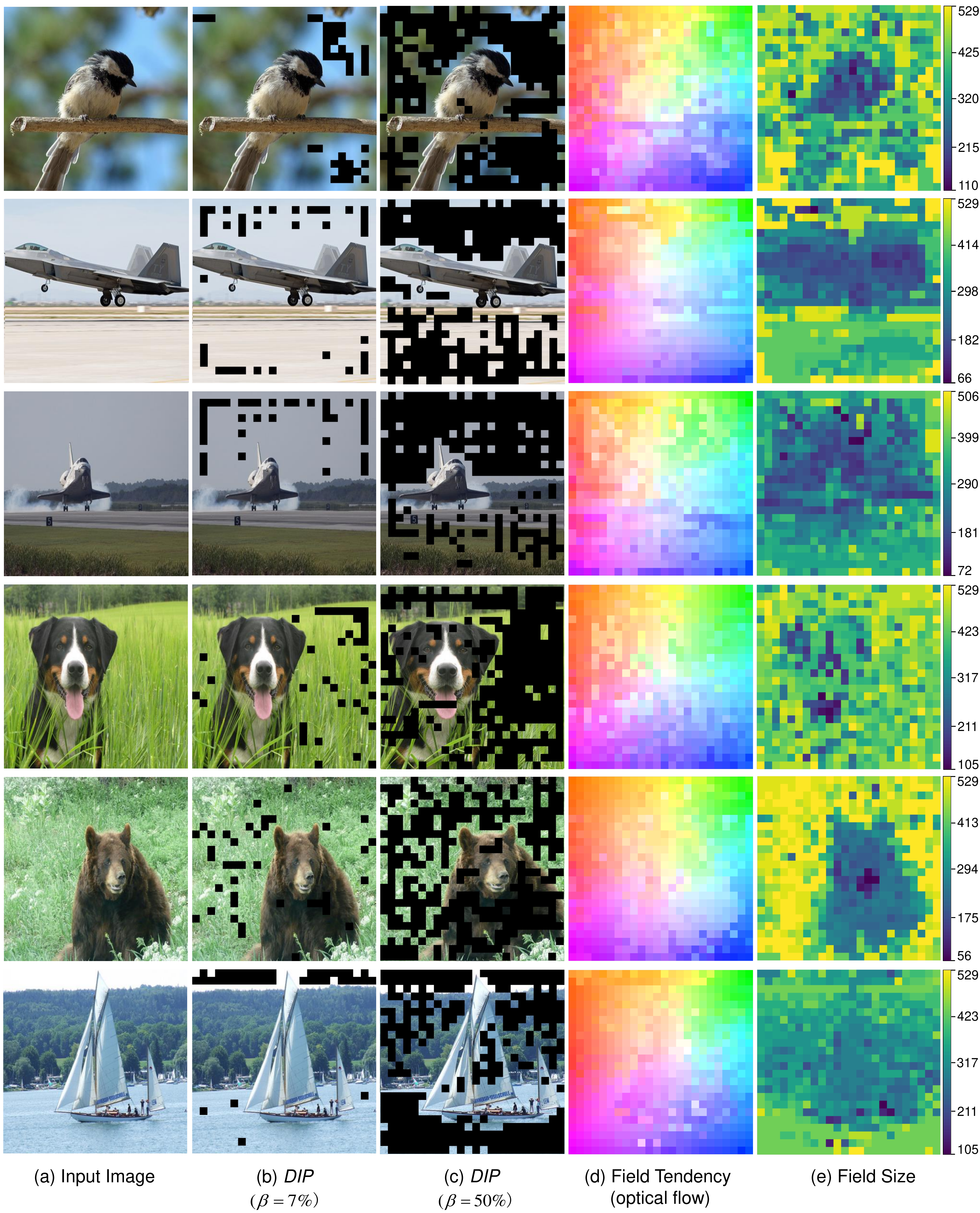}
	\label{sm_vis_area5}
\end{figure*}

\section*{4.Visualization of Predicted Attention Windows}
\begin{figure*}[!h]
\centering
		\includegraphics[width=0.9\linewidth]{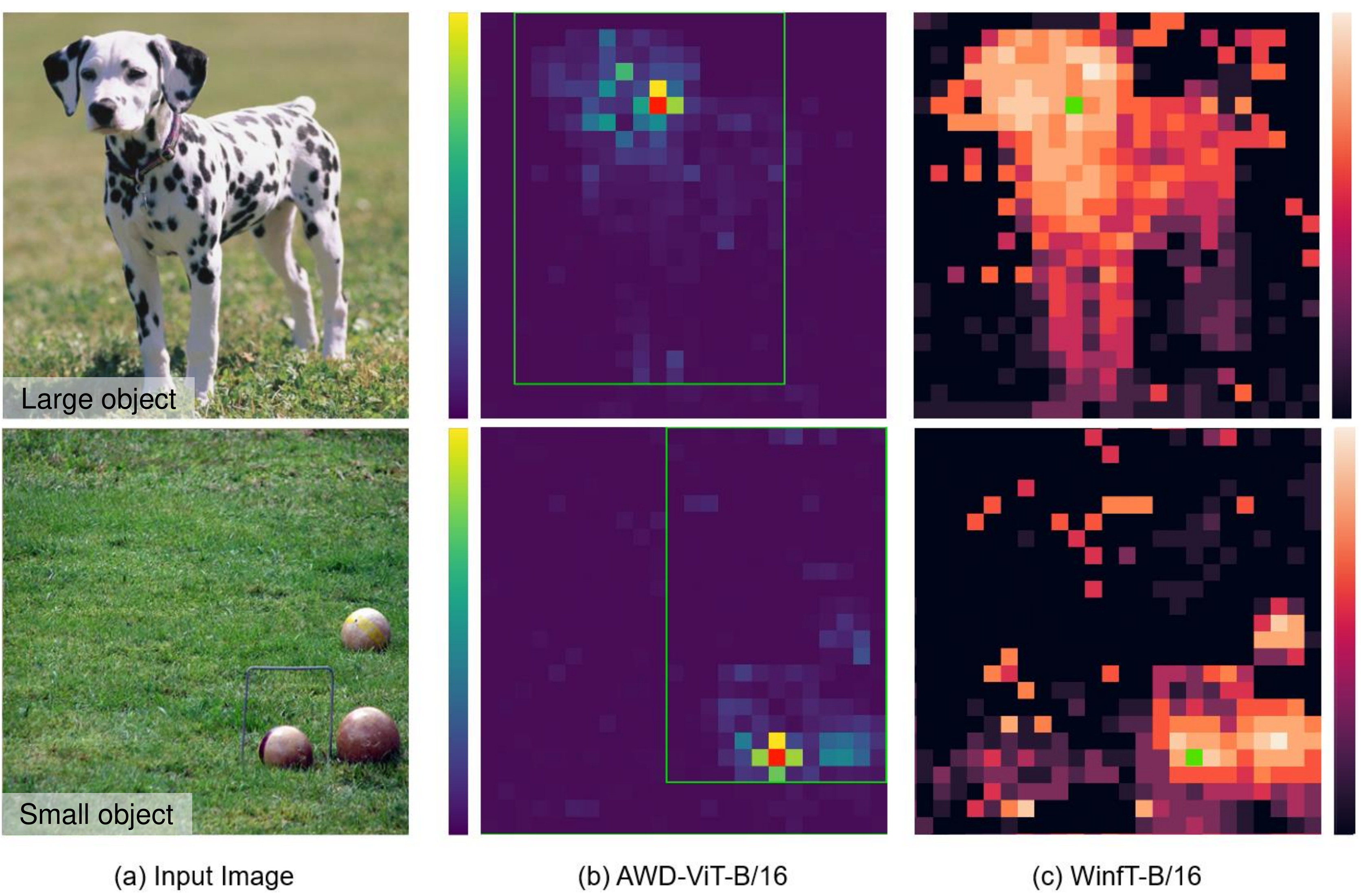}
	 \caption{Illustration of the responsive field from ViT (b) and the predicted attention mask in WinfT (c).}
	 \label{fig:predicted_mask}
\end{figure*}

We select two samples (large \& small objects) to visualize the ground-truth attention windows (b) and the predicted attention windows of WinfT (c) in \cref{fig:predicted_mask}. Since the predicted attention windows are different among all layers, we visualize the sum of predicted attention masks matrix $\sum_{l}^{12}W_{l}$ of window-free module across all 12 layers in WinfT (lighter block means more layers voting mask value of 1), where $W_l$ denotes the predicted binary mask in the $l$-th layer of WinfT. 

\end{document}